\documentclass[10pt,twocolumn,letterpaper]{article}
\usepackage{lipsum}
\usepackage{iccv}
\usepackage{times}
\usepackage{epsfig}
\usepackage{graphicx}
\usepackage{amsmath}
\usepackage{amssymb}
\usepackage{diagbox}
\usepackage{multirow}
\usepackage{makecell}
\usepackage[table ]{ xcolor}
\usepackage{booktabs}
\usepackage{makecell}
\usepackage[numbers]{natbib}
\usepackage{bbding}

\usepackage[breaklinks=true,bookmarks=false]{hyperref}

\iccvfinalcopy 

\def\iccvPaperID{****} 

\ificcvfinal\fi

\begin{document}

\title{Benchmarking and Analyzing Robust Point Cloud Recognition: \\ Bag of Tricks for Defending Adversarial Examples}

\author{Qiufan Ji$^1$, Lin Wang $^{2,3,*}$, Cong Shi$^1$, Shengshan Hu$^4$, Yingying Chen$^5$, Lichao Sun$^6$ \thanks{L. Wang and L. Sun are corresponding authors.}\\
$^1$New Jersey Institute of Technology, $^2$AI Thrust, HKUST(GZ), $^3$Dept. of CSE, HKUST, $^4$HUST, \\$^5$Rutgers University, $^6$Lehigh University \\
{\tt\small qj39@njit.edu, linwang@ust.hk, cong.shi@njit.edu, hushengshan@hust.edu.cn,}\\ {\tt\small yingche@scarletmail.rutgers.edu, james.lichao.sun@gmail.com}
}

\maketitle
\newcommand\blfootnote[1]{%
\begingroup
\renewcommand\thefootnote{}\footnote{#1}%
\addtocounter{footnote}{-1}%
\endgroup
}
\ificcvfinal\thispagestyle{empty}\fi

\begin{abstract}
     Deep Neural Networks (DNNs) for 3D point cloud recognition are vulnerable to adversarial examples, threatening their practical deployment. Despite the  many research endeavors have been made to tackle this issue in recent years, the diversity of adversarial examples on 3D point clouds makes them more challenging to defend against than those on 2D images. For examples, attackers can generate adversarial examples by adding, shifting, or removing points. Consequently, existing defense strategies are hard to counter unseen point cloud adversarial examples. In this paper, we first establish a comprehensive, and rigorous point cloud adversarial robustness benchmark to evaluate adversarial robustness, which can provide a detailed understanding of the effects of the defense and attack methods. We then collect existing defense tricks in point cloud adversarial defenses and then perform extensive and systematic experiments to identify an effective combination of these tricks. Furthermore, we propose a hybrid training augmentation methods that consider various types of point cloud adversarial examples to adversarial training, significantly improving the adversarial robustness. By combining these tricks, we construct a more robust defense framework achieving an average accuracy of 83.45\% against various attacks, demonstrating its capability to enabling robust learners. Our codebase are open-sourced on: \url{https://github.com/qiufan319/benchmark_pc_attack.git}.

     
\end{abstract}

\begin{figure}[htbp]
    \centering
    \includegraphics[width=0.5\textwidth]{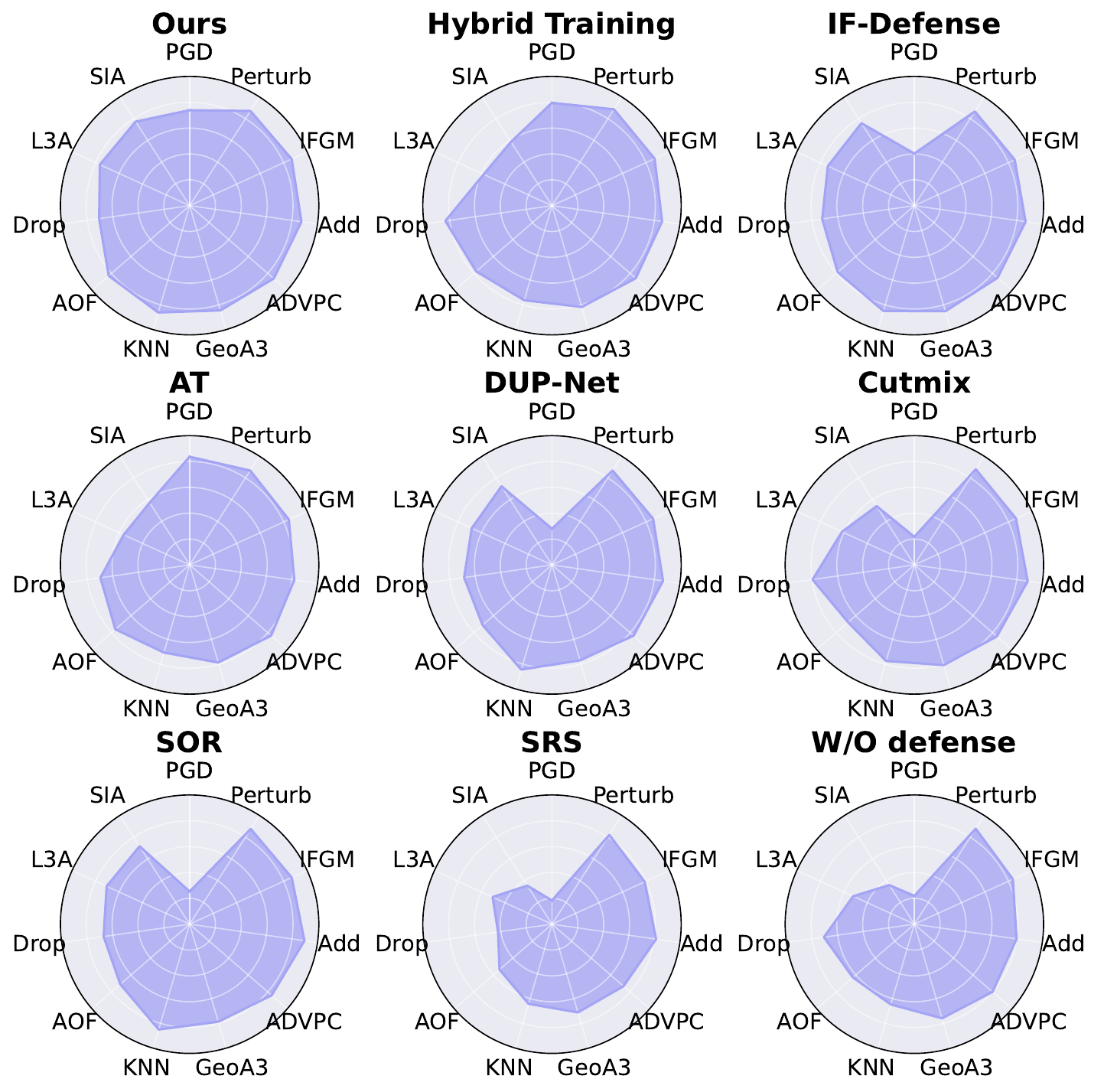}
    \caption{Point Cloud defense's adversarial robustness to various attacks in a radar chart. We evaluate the defense under 9 attack methods, including PGD~\cite{liu2019extending}, SIA~\cite{huang2022shape}, L3A~\cite{sun2021local}, Drop~\cite{zheng2019pointcloud}, AOF~\cite{liu2022boosting}, KNN~\cite{tsai2020robust},GeoA3~\cite{wen2020geometry}, ADVPC~\cite{hamdi2020advpc}, Add~\cite{xiang2019generating}, IFGM~\cite{liu2019extending}, Perturb~\cite{xiang2019generating}. Our method achieve good adversarial robustness against all attacks.}
    \label{fig:1}
    \vspace{-8pt}
\end{figure}

\vspace{-10pt}
\section{Introduction}
As an prominent form of 3D data representation, point clouds are extensively employed in various real-world sensing applications, such as autonomous driving~\cite{yue2018lidar}, robotics~\cite{pomerleau2015review}, and healthcare~\cite{aziz2016cloud}. To achieve precise perceive 3D objects, prior studies~\cite{qi2017pointnet,qi2017pointnet++,wang2019dynamic} have investigated the development of deep neural networks (DNNs) capable of detecting, segmenting, and identifying objects from point cloud data.
While these DNN-based methods have exhibited notable success, recent studies have exposed their susceptibility to adversarial examples~\cite{xiang2019generating,zheng2019pointcloud,wen2020geometry}. Specifically, the addition, removal, or shifting of a small proportion of 3D points from an object can significantly degrade the performance of the DNNs.

To mitigate the risk of adversarial examples, several defense strategies have been proposed to enhance the robustness of point cloud DNNs~\cite{wu2020if,zhou2019dup,liu2019extending}. For example, pre-processing techniques are applied to remove the points perturbed by adversarial examples~\cite{zhou2019dup,wu2020if}.
In addition, adversarial training~\cite{liu2019extending,tu2020physically,ma2021pointdrop}, which incorporates adversarial examples in the model training process, is designed to improve the adversarial robustness.    

Despite the initial success of investigating the adversarial robustness of point cloud DNNs, there are three obvious limitations for existing attacks and defenses:    

\textbf{L1: Unrealistic attack and defense scenarios.} 
    The current state-of-the-art (SOTA) adversarial learning has primarily focused on wihite-box attacks and defenses~\cite{xiang2019generating,wen2020geometry,zhou2019dup}, where the attacker has complete knowledge of the model architecture and paramteres. While these scenarios are useful for testing the limits of existing methods and understanding their vulnerabilities, they do not reflect the real-world security threat landscape. In many security-critical applications, such as autonomous driving and financial systems, attackers may not access to the model parameters.
 

\textbf{L2: Lack of a unified and comprehensive adversarial robustness benchmark.} While several studies~\cite{ren2022benchmarking,uy2019revisiting,taghanaki2020robustpointset,sun2022benchmarking} have been proposed to evaluate the robustness of point cloud DNNs, they have are all focused on benchmark under diverse types of corruptions. However, existing benchmarks research for studying adversarial robustness remains unexplored. Compared with the corruption-oriented attack methods, adversarial examples are difficult to be detected by both humans and machines. Moreover, perturbation generated using gradient descent are more effective than random corruptions, resulting in higher error rates and better imperceptibility. Despite recent studies exploring adversarial examples and defense on point cloud DNNs~\cite{xiang2019generating,tsai2020robust}, most of them has employ substantially different evaluation settings such as datasets, attacker's capability, perturbation budget, and evaluation metrics. The lack of a unified evaluation framework makes it challenging to fairly quantify the adversarial robustness. Additionally, current adversarial robustness evaluations only focus on one or a few attacks, defenses, and victim models, limiting the generalization and comparability of the results. For instance, the effectiveness of point cloud attack methods~\cite{xiang2019generating,wen2020geometry} is typically evaluated under a limited set of defenses and models. Similarly, defense strategies are often evaluated against only a few early attacks, making it difficult to capture their strengths and weaknesses based on incomplete evaluations.

\textbf{L3: Poor generalization of defense strategies.} Differ 2D image attack that modify the pixel value in a fixed data size, the adversarial example on point cloud offer a wider attack space and arbitrary data size. For instance, an attacker can generate adversarial example by adding, shifting, or removing points on the original point cloud. Unfortunately, most of existing defense strategies only consider one or two types, which can not handle unseen adversarial examples.

In this paper, we propose the first comprehensive and systematic point cloud adversarial robustness benchmark. Our benchmark provides a unified adversarial setting and comprehensive evaluation metrics that enable a fair comparison for both attacks and defenses. By analyzing the quantitative results, we propose a hybrid training strategy and construct a more robust defense framework by combining effective defense tricks. Our main contributions are summarized below:

\noindent\textbf{1) Pratical Scenario.} To evaluate the real-world performance of attacks and defenses, we refine the capability of both the attacker and defender, For example, we limited the maximum number of points added and the knowledge level of the attacker and defender. In our benchmark, all attackers are processed under the black-box setting, where the attacker does not have any additional knowledge about the model parameters, model structure, and training dataset.
\textbf{2) Unified Evaluation Pipeline.}
Our benchmarks provide a comprehensive and standardized evaluation methodology, enabling fair comparison and reproducibility of the proposed methods. For example, we evaluate the attack from attack success rate, transferability, and imperceptible, which are essential metrics for assessing the effectiveness, imperceptibility, and generalization of the attacks. 

\noindent\textbf{3) Bag-of-tricks for Defending Adversarial Examples.} Based on our adversarial robustness analyses with our benchmark, we proposed a hybrid training approach that jointly consider different types of adversarial examples, including adding, shifting, and removing points, to perform adversarial training. Through analysis of experiment result, we further construct a more robust defense framework by combining the effective defense tricks. As shown in Figure~\ref{fig:1}, our framework achieve the SOTA adversarial robustness under various attacks.


\section{Related works}
\label{p_2}
\noindent\textbf{3D Point Cloud Recognition.}
In contrast to 2D image data, 3D point cloud data is irregular and unordered, making it hard to be consumed by the neural networks designed for the 2D domain. PointNet~\cite{qi2017pointnet} is the pioneering work that directly consumes point cloud. It achieves permutation invariance of points by learning each point independently and subsequently using a symmetric function to aggregate features. Due to its high accuracy and efficiency, it has been widely used as the backbone for 3D point cloud recognition. As the update of PointNet, PointNet++~\cite{qi2017pointnet++} improves point cloud learning by capturing local information from the neighborhood of each point. Another representative work is DGCNN~\cite{wang2019dynamic}, which enhances the representation capability by building neighbor graphs between adjacent points and using a convolution-like operation (EdgeConv) on each connecting edge to capture local information. Recently, some transformer-based methods~\cite{muzahid2020curvenet,guo2021pct,xu2021learning} have been proposed, achieving good performance.

\noindent\textbf{Robustness Benchmark for Point Cloud.}
Several benchmarks ~\cite{sun2022adversarial,sun2020adv,cao2303comprehensive,sun2023scalable,hu2023pointca} have been built for studying the robustness of point cloud learning. ~\cite{reizenstein2021common} build a real-world dataset to evaluate the gap between simulation and real-world. To evaluate the corruption robustness, ModelNet-C~\cite{sun2022benchmarking} categorizes common corruptions and builds a new corruption dataset ModelNet-C by corrupting the ModelNet40 test set with simulated corruptions like occlusion, scale, and rotation. RobustPointset~\cite{taghanaki2020robustpointset} evaluates the robustness of point cloud DNNs and shows that existing data augmentation methods can not work well to unseen corruptions. However, little attention has been paid to adversarial examples of point cloud recognition. In this paper, we aim to present the first comprehensive, systematic benchmark to evaluate the point cloud adversarial examples and defenses.

 \begin{figure*}[t]
    \centering
    \includegraphics[width=0.90\textwidth]{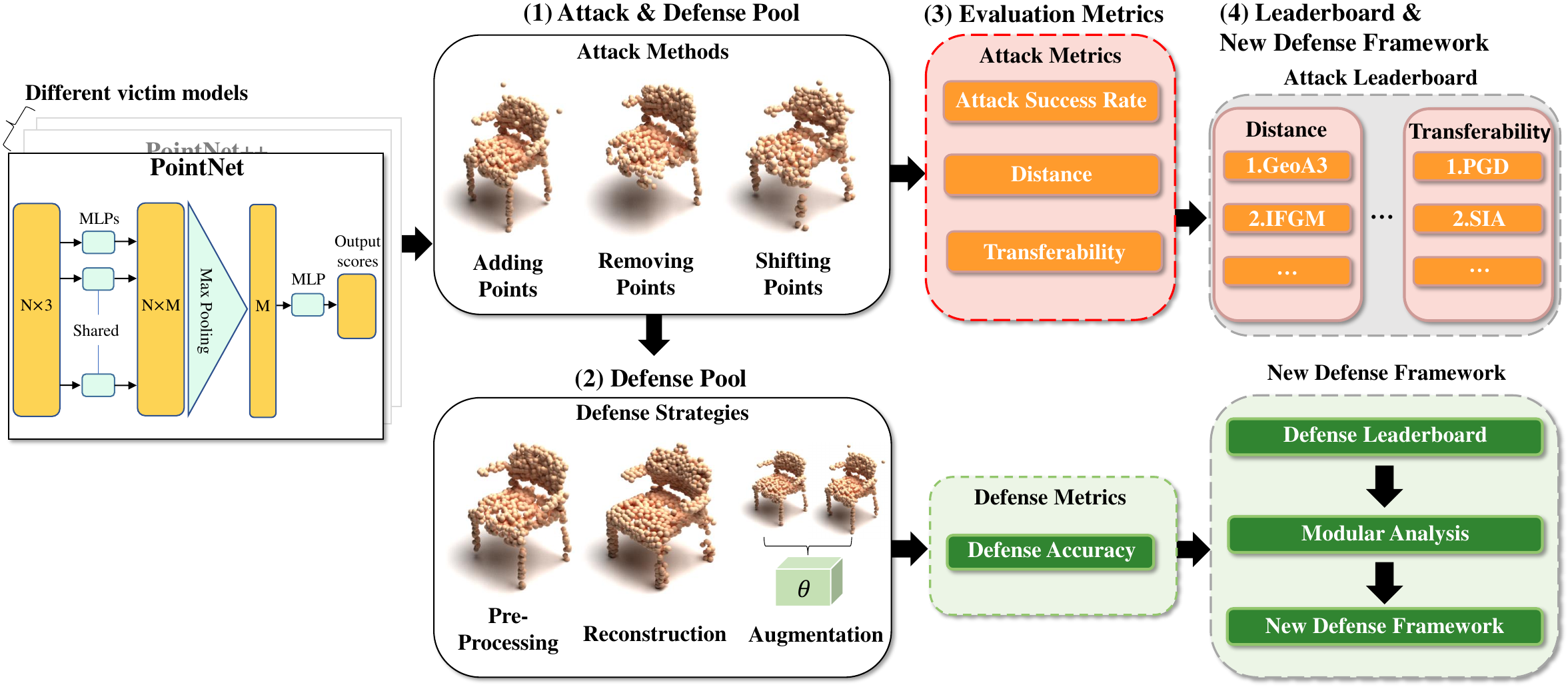}
    \caption{The pipeline of our benchmark. }
    \label{fig:2}
    \vspace{-10pt}
\end{figure*}

\section{Benchmark}
\label{p_3}
\subsection{Preliminaries}
\noindent\textbf{Problem Formulation.} We defined the point cloud as $X\in \mathbb{R}^{N\times3}$, where $N$ is the number of points . Each point $x_i\in \mathbb{R}^3$ indicates the 3D coordinate $(x_i,y_i,z_i)$. Formally, a classifier $f_{\theta}(X)\rightarrow Y$ maps the input point cloud $X$ to its corresponding label $y\in Y$ with parameter $\theta$. For adversarial examples on point cloud DNNs, an attacker generates an adversarial example $\hat{X}$, which makes the classifier $f_{\theta}$ output an incorrect label $\hat{Y}$. Generally, the objective function of generating adversarial examples can be formulated as:
\begin{equation}
\label{equation 1}
\begin{split}
    \mathop{\min} D(X,\hat{X}), \qquad \mathrm{ s.t.} \ f_{\theta}(\hat{X})={\hat{Y}},
\end{split}
\end{equation}
where $D(\cdot,\cdot)$ is the distance function measuring similarity between $X$ and $\hat{X}$. The distance is normally constrained to a small budget $\rho$ ensuring the imperceptibility. Because the equation (\ref{equation 1}) is non-convex, according to ~\cite{xiang2019generating} we reformulated it as gradient-based optimization algorithms:
\begin{equation}
    \mathop{\min}{f_{adv}(X,\hat{X})+\lambda * D(X,\hat{X})} \qquad \mathrm{ s.t.}\ D(X,\hat{X})<\rho,
\end{equation}
where $f_{adv}$ is the adversarial loss function, including logits loss and cross-entropy loss, and $\lambda$ is a hyperparameter to balance distance and adversarial loss.


\noindent\textbf{Attack Types.} An attacker can have different targets of generating adversarial examples. In our benchmark, we divided the attacks into targeted and untargeted. Targeted: A targeted attack tries to make the victim model outputs a result that it desired, as $f_{\theta}(\hat{X})=Y^*$, where $y^*$ is the target label. Untargeted: an untargeted attack only aims to make the victim model outputs a wrong result, as $f_{\theta}(\hat{X})\neq Y$, where $Y$ is the true label.

\noindent\textbf{Attack Knowledge.} The attacker can have different levels of knowledge of the victim model. Based on the knowledge level, the attacks can be divided into Black-Box and White-Box. Black-Box: The attacker can not get any information about the victim model, such as gradient information, model structure, and parameters. However, they have limited chances to query the victim model and obtain the output. White-Box: The attacker can get any information about the victim model. In both knowledge settings, the attacker can access the training dataset. 

\noindent\textbf{Attack Stage.} Based on the stage where the attacks happened, we divided the attacks into Poisoning and Evasion. Poisoning: The attacker generate the adversarial examples and inject them into the training dataset. Once the attacker change the training dataset, the victim model will be retrained on changed dataset to get a worse model. Evasion: The parameter of the victim model is fixed, and attackers inject adversarial perturbation into testing data.



\subsection{Practice Scenario}
In real-world, the victim model is usually trained in a confidential manner, and the attacker is hard to modified the model meaning that  white-box and poisoning setting are normally infeasible. 

In our benchmark, we make the following assumptions for a unified and practical adversarial robustness evaluation protocol: (1) Black-box: the attacker does not know the defender's strategies, and vice versa. (2) Evasion: The point cloud DNNs are trained with trusted data and training model is inaccessible to the attacker. (3) Untargeted: in our benchmark, we select untargeted attacks for the evaluation of adversarial robustness. Because untargeted attack is easier than targeted attacks for attacker, thus the untargeted attack is the upper bound of attack intensities and more difficult for defense strategies.  
We define the full capabilities of attackers and defenders in our benchmark:

\noindent\textbf{Attacker}: 1) The attacker can access the testing point cloud data to produce adversarial examples, but they should not have knowledge about the victim model or defense mechanism. 2) To preserve adversarial examples imperceptible, the attacker is only allowed to add or delete a limited number of points in the point cloud. 3) The attacker can not modify the training dataset. 4) The attacker can only query the victim model with limited times.

\noindent\textbf{Defender}: 1) The defender has full access to the training dataset. 2) The defender can use any solution to improve the robustness without additional knowledge about the attack.

\noindent\textbf{Both sides}: We assume that attackers know the architecture of victim model (e.g., PointNet, PointNet++), and then they can train a corresponding surrogate model to generate adversarial examples. Similarly, the defender can have some assumptions on the effects of adversarial examples (e.g., point cloud adversarial examples usually exist some outliers). For both the attacker and the defender, the generalization (e.g., an attack can bypass multiple defense techniques) is an important factor for adversarial robustness quantification. Thus, we evaluate the effectiveness of SOTA attacks against various defense techniques and model architecutres. We also conduct similar quantifications for the defenses in our benchmark.

By following the above rules, we provide a unified evaluation scenario for attacks and defenses in a principled way. It is worth nothing that the unified scenario is not the only valid ones, our benchmark will include more scenarios as this field evolved over time.
As shown in Figure \ref{fig:2}, the attack and defense pool include all attack methods and defense strategies in our benchmark. Our evaluation metrics incorporate three attack metrics, namely, attack success rate, distance, and transferability, to assess the performance of attack methods. Additionally, we use one defense accuracy metric to evaluate the effectiveness. We construct attack and defense leaderboards based on the metrics values. Further, we conduct modular analysis on each defense strategy, and subsequently integrate effective modules to construct a more robust defense framework.


\subsection{Generating Adversarial Examples}
\label{p_4}
Adversarial examples were first discovered by ~\cite{szegedy2013intriguing} in 2D image classification tasks. With the development of adversarial learning, some works ~\cite{xiang2019generating,huang2022shape,wen2020geometry} proved that point clouds also be vulnerable to adversarial examples. The adversarial examples on point cloud can be divided into adding points, removing points, and shifting points attacks.

\noindent\textbf{Adding Points.} The attacker generate adversarial examples by adding a set of adversarial points $Z\in \mathbb{R}^{k\times3}$ where $k$ is the number of modified points in different attack settings.
Given the adversarial perturbations $\rho \in \mathbb{R}^{k\times3}$ on added points, the objective function of adding points attacks can be formulated as:
\begin{equation}
    \mathop{\min}{f_{adv}(X,X\cup(Z+\rho))-\lambda D(X,X\cup(Z+\rho))},
\end{equation}
Adding independent points~\cite{xiang2019generating} chooses the critical points that are still active after max-pooling operation, as the initialized positions, and then uses C\&W~\cite{carlini2017towards} to output their final coordinates. Although other adding points attacks exist, such as add clusters~\cite{xiang2019generating} and adversarial sticks~\cite{liu2020adversarial}. These methods are not practical because they create a noticeable continuous deformation and then produce large perturbations. Consequently, for the purpose of adding points attack, only independent points are considered.

\noindent\textbf{Removing Points.} The attacker remove some points to spoof the classifier. As the representative work of removing points attack, saliency map~\cite{zheng2019pointcloud} constructs a saliency map to measure the contribution of each point and then removes the points based on the saliency score. In our benchmark, we limit the number of dropped points to keep the drop attack imperceptible.

\noindent\textbf{Shifting Points.} The attacker perturbs the coordinates of a set of points to implement an attack. The objective function of shifting points attacks can be formulated as:
\begin{equation}
        \mathop{\min}{f_{adv}(X,(X+\rho))-\lambda D(X,(X+\rho))},
\end{equation}
The iterative fast gradient method (IFGM)~\cite{liu2019extending} is an extension of the fast gradient method (FGSM) that repeats FGSM multiple times to generate better adversarial examples. The project gradient descent method (PGD) ~\cite{liu2019extending} projects the perturbed point onto the triangle plane where the points are sampled. Perturb~\cite{xiang2019generating} proposes a C\&W based algorithm to generate adversarial examples. To reduce the outliers, KNN~\cite{tsai2020robust} incorporates Chamfer measurement and KNN distance to encourages the compactness of local neighborhoods in the generated adversarial examples. GeoA3~\cite{wen2020geometry} perturbs points in a geometrically aware way by adding local curvatures to the loss function, thereby making the adversarial examples more imperceptible. L3A~\cite{sun2021local} proposes a novel optimization method to avoid local optima, making the attack more efficient. AdvPC~\cite{hamdi2020advpc} utilizes a point cloud auto-encoder (AE) during the generation, improving the transferability of adversarial examples. SIA~\cite{huang2022shape} builds a tangent plane to each point and limits the point perturbation along the optimal direction on the tangent plane, making the adversarial examples more imperceptible. AOF~\cite{liu2022boosting} proposes a more robust and transferable attack by perturbing the low-frequency in the frequency domain.

\section{Analysis and Bag-of-Tricks for Defending Adversarial Examples}
\begin{figure}[t]
    \centering
    \includegraphics[width=0.5\textwidth]{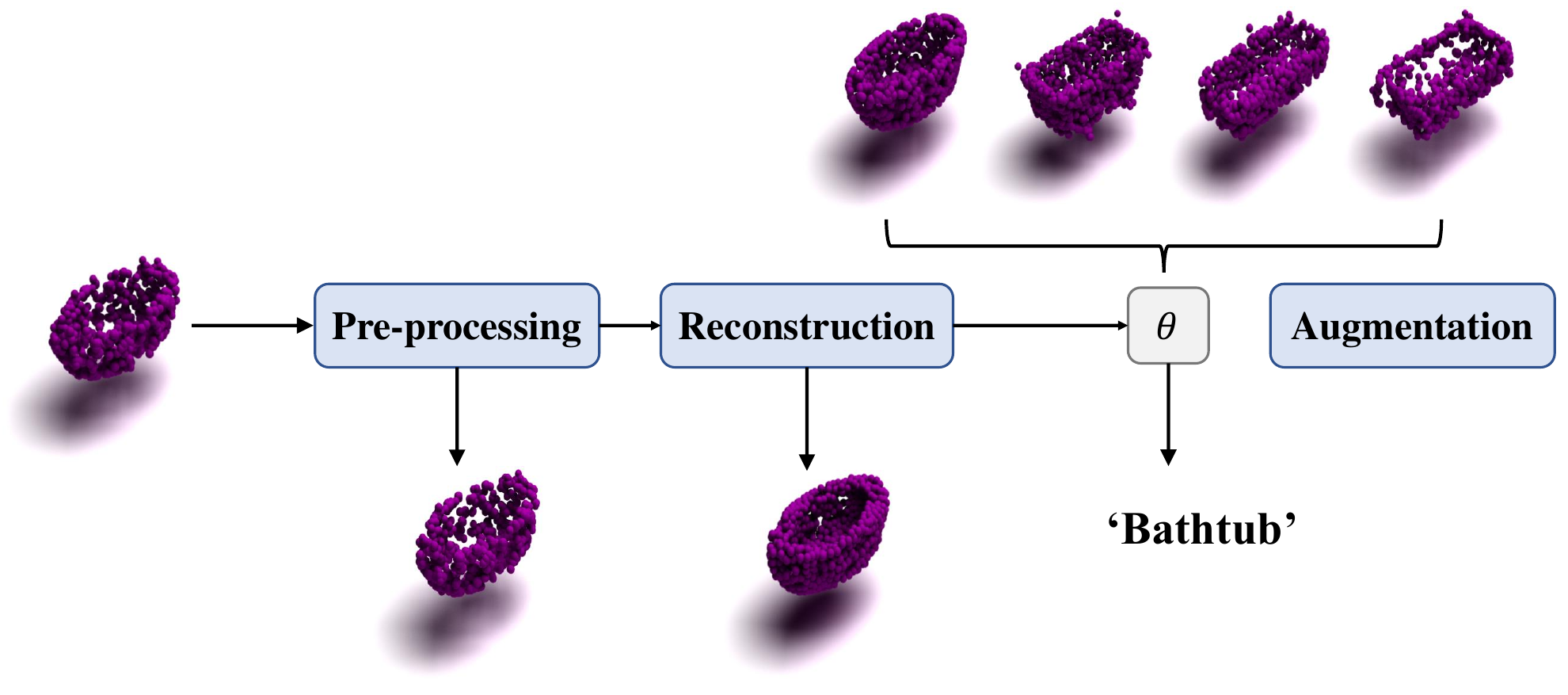}
        \caption{Robust defense framework paradigm. The adversarial robustness of point clouds against various attacks is influenced by three critical components, including pre-processing, reconstruction, and augmentation methods.}
    \label{fig:3}
    \vspace{-5pt}
\end{figure}

To alleviate the adversarial behaviors, the most popular defending techniques can be divided into three directions, i.e., pre-processing, reconstruction, and augmentation, as shown on Figure~\ref{fig:3}.

\noindent\textbf{Pre-processing.} Advanced pre-processing aims to reduce the noise before inference. A straightforward approach is Simple Random Sampling (SRS), which random samples a subset of points from the original point cloud as input. Statistical Outlier Removal (SOR)~\cite{zhou2019dup} computes KNN distance and removes the points with a large KNN distance. 

\noindent\textbf{Reconstruction.} Adversarial examples often result in the absence or alteration of geometric features in the original point cloud. With the development of 3D point cloud reconstruction, some works employed 3D reconstruction networks to improve robustness. We consider two 3D Reconstruction networks in our benchmark:

\textbf{DUP-Net}~\cite{zhou2019dup}: DUP-Net employs the PU-Net~\cite{yu2018pu} as its reconstruction network. The PU-Net utilizes point cloud up-sampling to reconstruct the surface and generate a high-resolution point cloud that captures the missing structures of the object's surface. More experiment results of DUP-Net can be found in the appendix.

\textbf{IF-Defense}~\cite{wu2020if}: In contrast to DUP-Net, IF-Defense employs the ConvONet~\cite{wu2019pointconv} as its reconstruction network. The ConvONet uses the point cloud as input for shape latent code extraction, while the encoder produces a learned implicit function network. By pre-training the implicit model on clean data, the decoder's output space situates on the accurate and complete shape manifold.

We present the results of the adversarial robustness evaluation of IF-Defense and ConvONet in the appendix. We find that both reconstruction networks can improve adversarial robustness. Especially, ConvONet, with its superior 3D reconstruction performance, exhibits better adversarial robustness in most attacks.

\noindent\textbf{Augmentation.} The principle of augmentation is aimed at enhancing the robustness of the model when encountering minor noise. One notable approach is adversarial training~\cite{liu2019extending}, which incorporates adversarial examples into the training phase. Another augmentation method is PointCutmix~\cite{zhang2021pointcutmix}, which utilizes mix-based point cloud to enhance the model's robustness. However, due to the variety of attack types, existing augmentation methods performance poorly against adversarial attacks in point cloud. 

\textbf{Hybrid Training.} To enhance the performance of augmentation, we propose a hybrid training method that leverages multiple attack approaches. Especially, hybrid training selects $k$ attack approaches, including adding, removing, and shifting attacks. For each class in the dataset, the proposed method divides the samples equally into $k$ parts and applies different attack approaches to each part. Finally, all generated adversarial examples are integrated to augment the training data. The result of adversarial robustness of augmentation methods is reported in the appendix. Our hybrid training achieves the highest level of adversarial robustness among all augmentation methods.

In Table~\ref{table::AT}, we show the accuracy of defense strategies. We find the three components can improve the robustness of adversarial robustness. Based on the aforementioned analyses, we propose a robust defense framework that integrates SOR, hybrid training, and ConvONet. In Table~\ref{table::summary_of_defense}, our defense framework demonstrates superior robustness compared to other existing defense strategies. Moreover, in the ablation study presented in the appendix, we demonstrate that all aforementioned modules contribute significantly to the adversarial robustness of our defense framework, with hybrid training being the critical component for enhancing adversarial robustness. These results substantiate our conclusions from the modular analysis and establish our framework as a solid baseline for future research on adversarial robustness.

\begin{table}[t]
\centering
\caption{The accuracy of defense strategies. }
\label{table::AT}
\resizebox{1\linewidth}{!}{
\begin{tabular}{l|c c c}
\toprule[1pt]
 & AOF & GEOA3 & SIA \\
\midrule
w/o defense & 54.54 & 61.26 &31.40\\
SOR (Pre-processing) & 68.48 & 73.22 &59.12
\\
IF-Defense (Reconstruction) & 66.99 & 65.68 &43.76\\
Hybrid Training (Augmentation) & 73.43 & 75.45 &76.26\\
\bottomrule[1pt]
\end{tabular}}
\vspace{-15pt}
\end{table}

\begin{table*}[t]
\centering
\caption{Leaderboard. Bold: best in column. Underline: second best in column. Blue: best in row. Red: worst in row. Compared with existing augmentation methods, our hybrid training achieves the SOTA performance. }
\small
\label{table::summary_of_defense}
\resizebox{0.9\linewidth}{!}{
\begin{tabular}{p{2.5cm}<{\centering} p{1.5cm}<{\centering}|p{1cm}<{\centering} p{0.75cm}<{\centering} p{0.75cm}<{\centering} p{0.75cm}<{\centering} p{0.75cm}<{\centering} p{0.75cm}<{\centering} p{0.75cm}<{\centering} p{0.75cm}<{\centering} p{0.75cm}<{\centering} p{0.75cm}<{\centering} p{0.75cm}<{\centering} p{0.75cm}<{\centering}}
\toprule[1pt]
    Defense \& (Acc) & Model & Clean & PGD & SIA & L3A & Drop & AOF& KNN &  GeoA3 & AdvPC & Add & IFGM & Perturb \\ 
\midrule
\multirow{8}*{\makecell[c]{{Ours} \\(\textbf{83.45})}}
&  PointNet &87.36	&76.70	&76.26	&\cellcolor{red!10}{75.04}	&80.79	&81.60	&85.53	&84.04	&86.22	&\cellcolor{blue!10}{87.28}	&86.26	&86.35
\\
~&PointNet++& 88.65	&\cellcolor{red!10}{71.56}	&78.61	&76.70	&83.14	&\underline{84.12}	&86.87	&85.78	&\textbf{87.40}	&88.45	&88.17	&\cellcolor{blue!10}{88.70}
\\
~&DGCNN &87.97	&\cellcolor{red!10}{72.93}	&\underline{81.81}	&77.96	&82.94	&84.08	&86.39	&84.81	&86.71	&\cellcolor{blue!10}{88.17}	&87.76	&87.32
\\
~&Pointconv &87.12	&\cellcolor{red!10}{70.14}	&80.67	&76.50	&83.79 &81.77	&86.30	&85.53	&86.83	&\cellcolor{blue!10}{87.88}	&87.52	&86.95
\\
~&PCT &83.27	&\cellcolor{red!10}{73.91}	&78.32	&75.77	&76.86	&79.66	&82.41	&80.71	&81.60	&82.78	&82.21	&\cellcolor{blue!10}{83.71}
\\
~&Curvenet &88.57	&\cellcolor{red!10}{76.13}	&80.26	&\underline{78.16}	&83.67	&83.59	&\underline{87.20}	&84.97	&86.14	&\cellcolor{blue!10}{88.05}	&86.99	&87.88
\\
~&RPC & 88.86	&\cellcolor{red!10}{74.11}	&\textbf{82.66}	& \textbf{78.20}	&82.98&	\textbf{84.48}	&\textbf{87.93}	&85.49	&\underline{87.38}	&88.01	&\cellcolor{blue!10}{88.45}	&88.13
\\
~&GDANet & 88.57	& \cellcolor{red!10}{75.32}	& 80.79 & 	77.31	& 83.59	&83.75	& 86.59	& 84.44	& 86.63	& \cellcolor{blue!10}{88.65}	& 86.95	& 87.28
\\
\midrule
\midrule

\multirow{8}*{\makecell[c]{Hybrid Training \\ (79.35)}}
&  PointNet &88.57	& \underline{80.15}	&53.08	&\cellcolor{red!10}{50.28}	&77.55	&73.34	&64.71	&75.45	&84.12	&83.39	&85.17	&\cellcolor{blue!10}{87.64}
\\
~&PointNet++& 89.75	&77.39	&\cellcolor{red!10}{52.80}	&57.74	&85.74	&79.85&	74.68	&82.74	&86.08	&85.45	&88.29	&\cellcolor{blue!10}{89.00}
\\
~&DGCNN &89.47	&\textbf{81.40}	&66.29	&\cellcolor{red!10}{61.14}	&\underline{86.14}	&80.19	&80.59	&82.74	&87.28	&87.76	&\underline{88.53}	&\cellcolor{blue!10}{\underline{89.95}}
\\
~&Pointconv &90.19	&80.06	&\cellcolor{red!10}{45.30}	&57.37	&\textbf{86.47}	&69.65	&81.69	&83.31	&85.13	&88.82	&\textbf{89.83}	&\cellcolor{blue!10}{\textbf{90.28}}
\\
~&PCT &87.44	&\cellcolor{red!10}{45.95}	&75.97	&76.70	&72.69	&78.57	&85.86	&83.02	&86.35&	87.12	&\cellcolor{blue!10}{87.86}	&87.24
\\
~&Curvenet &87.16	&\cellcolor{red!10}{44.57}	&76.22	&76.00	&74.15	&81.48	&85.45	&83.43	&86.14	&87.40	&\cellcolor{blue!10}{87.84}	&86.35
\\
~&RPC & 85.45	&\cellcolor{red!10}{53.85}	&76.46	&74.80	&70.30	&80.31	&83.95	&82.17	&84.60	&84.52	&\cellcolor{blue!10}{85.90}	&84.68
\\
~&GDANet & 87.24	&\cellcolor{red!10}{38.74}	&79.01	&75.04	&72.16	&80.71	&85.41	&82.86	&85.21	&86.43	&\cellcolor{blue!10}{86.87}	&86.67\\
\midrule
\midrule

\multirow{8}*{ \makecell[c]{IF-Defense\\ (78.4)}}
&  PointNet & 85.33	&\cellcolor{red!10}{44.89}	&68.60	&68.76	&65.19	&75.28	&82.46	&82.74	&83.79	&\cellcolor{blue!10}{85.41}	&82.86	&85.01
\\
~&PointNet++ &87.52	&\cellcolor{red!10}{38.61}	&72.45	&72.33	&73.01	&78.28	&85.56	&85.17	&85.66	&\cellcolor{blue!10}{87.20}	&85.37	&87.12
\\
~&DGCNN &87.88	&\cellcolor{red!10}{40.32}	&77.92	&76.05	&71.48	&80.35	&85.78	&85.41	&85.49	&86.75&	85.86	&\cellcolor{blue!10}{87.32}
\\
~&Pointconv &85.53	&\cellcolor{red!10}{28.61}	&78.24	&73.66	&73.87	&75.89	&84.85	&84.24	&84.48	&\cellcolor{blue!10}{85.58}	&84.20	&85.37
\\
~&PCT & 88.33	& \cellcolor{red!10}{45.83}	& 75.24	& 73.45	& 72.85	& 79.42	& 85.86	& 85.29	& 86.35	& \cellcolor{blue!10}{87.16}	& 85.94	& 86.83
\\
~&Curvenet &88.33	&\cellcolor{red!10}{45.38}	&76.18	&75.45	&74.19	&80.51	&85.45	&\underline{86.02}	&86.14	&\cellcolor{blue!10}{88.01}	&86.75	&86.67
\\
~&RPC &88.05	&\cellcolor{red!10}{40.44}	&76.13	&73.62	&73.58	&77.43	&83.95	&\textbf{86.06}	&84.60	&\cellcolor{blue!10}{87.72}	&85.90	&87.64
\\
~&GDANet &87.93	&\cellcolor{red!10}{38.05}	&81.65	&75.45	&72.37	&80.92	&85.41	&85.94	&85.21	&\cellcolor{blue!10}{87.72}	&86.10	&86.87
\\
\midrule
\midrule

\multirow{8}*{\makecell[c]{SOR\\(75.19)}}
&  PointNet &86.95	&\cellcolor{red!10}{42.10}	&63.21	&63.70	&57.86	&68.48	&80.06	&73.22	&80.49	&\cellcolor{blue!10}{86.10}	&84.16	&85.53
\\
~&PointNet++ &88.57	&\cellcolor{red!10}{25.00}	&64.30	&72.49	&66.25	&71.31	&85.13	&80.23	&84.89	&88.70	&87.72	&\cellcolor{blue!10}{88.98}
\\
~&DGCNN& 88.57	&\cellcolor{red!10}{18.00}	&73.58	&69.89	&66.94	&66.25	&85.25	&74.24	&82.33	&\cellcolor{blue!10}{87.88}	&87.64	&87.44
\\
~&Pointconv &72.12	&\cellcolor{red!10}{11.70}	&71.84	&69.73	&72.12	&65.92	&85.53	&77.47	&84.68	&\cellcolor{blue!10}{88.49}	&86.02	&87.12
\\
~&PCT & 88.41	&\cellcolor{red!10}{38.94}	&72.97	&70.75	&67.50	&74.84	&85.01	&80.31	&84.52	&\cellcolor{blue!10}{88.65}	&86.79	&87.76
\\
~&Curvenet &88.33	&\cellcolor{red!10}{33.63}	&74.51	&76.86	&69.81	&76.62	&86.43	&83.39	&85.17	&\cellcolor{blue!10}{\textbf{89.10}}	&86.83	&87.72
\\
~&RPC & 89.43	&\cellcolor{red!10}{15.07}	&72.57	&69.17	&69.00	&69.85	&85.53	&78.04	&84.04	&\cellcolor{blue!10}{\underline{88.86}}	&87.44	&87.72
\\
~&GDANet & 89.26 	&\cellcolor{red!10}{17.34}	&79.90	&72.12	&68.40	&74.39	&86.59	&82.29	&85.94	&\cellcolor{blue!10}{88.85}	&88.01	&87.88
\\
\midrule
\midrule

\multirow{8}*{\makecell[c]{No Defense\\(67.06)}}
&  PointNet & 87.64	&34.32	&\cellcolor{red!10}{31.40}	&45.38	&59.64	&54.54	&45.10	&61.26	&76.94	&71.76	&74.59	&\cellcolor{blue!10}{85.58}
\\
~&PointNet++ &89.30	&\cellcolor{red!10}{15.56}	&16.82	&44.89	&71.47	&60.01	&54.25	&74.51	&73.62	&72.37	&81.22	&\cellcolor{blue!10}{88.17}
\\
~&DGCNN & 89.38	&\cellcolor{red!10}{18.96}	&51.01	&57.25	&73.10	&62.84	&70.10	&77.39	&76.86	&83.71	&86.91	&\cellcolor{blue!10}{88.74}
\\
~&Pointconv &88.65	&\cellcolor{red!10}{9.81}	&25.41	&47.57	&76.50	&51.26	&71.80	&77.67	&76.90	&85.15	&86.51	&\cellcolor{blue!10}{88.09}
\\
~&PCT &89.99	&\cellcolor{red!10}{32.33}	&41.05	&53.75	&71.27	&65.92	&67.08	&78.32	&82.58	&82.33	&85.62	&\cellcolor{blue!10}{89.22}
\\
~&Curvenet &89.47	&\cellcolor{red!10}{27.96}	&38.70	&53.53	&71.29	&69.52&	66.73	&79.17	&84.48	&79.86	&85.09	&\cellcolor{blue!10}{88.33}
\\
~&RPC&89.42	&\cellcolor{red!10}{15.36}	&32.33	&54.01	&69.89	&72.16	&70.58	&77.92&	83.67	&80.75	&83.55	&\cellcolor{blue!10}{85.98}
\\
~&GDANet& 89.10 	& \cellcolor{red!10}{20.71}	& 50.57	& 59.64	& 72.33	& 67.30	& 72.20	& 80.92	& 85.13	& 83.47	& 86.63	& \cellcolor{blue!10}{88.33}
\\
\midrule
 \makecell[c]{Avg.ASR} &  &  -& 51.84 & 36.15 & 33.85 &27.19 &25.41 &20.59 &18.83 &15.77 &14.26&13.98 &12.49 \\
\bottomrule[1pt]
\end{tabular}}
\end{table*}
\section{Experiments}
\label{p_5}
\subsection{Experimental Setup}
\noindent\textbf{Dataset and DNNs.} All of our experiments are conducted commonly on ModelNet40 dataset. ModelNet40 consists of 123,11 CAD models for 40 object classes. In our experiments, we split ModelNet40 dataset into two parts: 9,843 and 2,468 samples for training and testing, respectively. Following ~\cite{qi2017pointnet++}, we use farthest points sampling (FPS) to uniformly sample 1024 points from the surface of each object as input data. We adopt eight widely used point cloud DNNs as victim models, including PointNet~\cite{qi2017pointnet}, Pointnet++~\cite{qi2017pointnet++}, DGCNN~\cite{wang2019dynamic}, PointConv~\cite{wu2019pointconv}, PCT~\cite{guo2021pct}, Curvenet~\cite{muzahid2020curvenet}, PRC~\cite{ren2022benchmarking}, and GDANet~\cite{xu2021learning}. For PointNet++ and PointConv, we use the single scale grouping (SSG) strategy. All models are trained without data augmentation.

\noindent\textbf{Attack Settings.} According to the attacker capability setting, we implemented all attacks on the testing dataset and using a PointNet model as surrogate model. It should be note that the hyperparameters of the surrogate model differed from those of the victim models. Specifically, We employed 11 different attacks. In the adding points attack~\cite{xiang2019generating}, we added 100 points, and in the removing points attack~\cite{zheng2019pointcloud}, we removed 200 points. Regarding the shifting points attack, we utilized a range of methods, including SIA~\cite{huang2022shape}, L3A~\cite{sun2021local}, KNN~\cite{tsai2020robust}, GeoA3~\cite{wen2020geometry}, IFGM~\cite{liu2019extending}, PGD~\cite{liu2019extending}, Perturb~\cite{xiang2019generating}, AOF~\cite{liu2022boosting} and AdvPC~\cite{hamdi2020advpc}. To ensure fair verification, we constrained all Shifting points adversarial examples equally using an $l_{\infty}-$ normal ball with a radius of 0.16, and we performed untargeted attacks under the same setting.

\noindent\textbf{Defense Settings.} For SRS~\cite{zhou2019dup}, we randomly dropped 500 points from the input point cloud. To perform SOR~\cite{zhou2019dup}, we first computed the average distance from its $k$-nearest neighbors and subsequently removed points if the average distance exceeded the threshold of $\mu+\alpha \cdot \sigma$, where $\mu$ and $\sigma$ are the mean and standard deviation, respectively, and $k$ and $\alpha$ are hyperparameters. We set the hyperparameters to be $k=2$ and $\alpha=1.1$. For IF-Defense~\cite{wu2020if}, we chose ConvOnet~\cite{peng2020convolutional}, which achieved the superior performance for most attacks in their experiment results. In adversarial training, all victim models were trained on clean data and adversarial examples generated by PGD with $l_{\infty}=0.20$. For hybrid training, we combined adding independent points, saliency map, and PGD, with adversarial training. In the adding independent points attack, we added 200 points to point cloud. In the saliency map attack, we removed 300 points form the point cloud based on their saliency map. In PGD, we set the perturbation budget to $l_{\infty}=0.20$.

\noindent\textbf{Evaluate Metrics.} To evaluate the imperceptibility of generated adversarial examples, we adopt Chamfer Distance (CD) and Hausdorff Distance (HD) as distance metrics for each adversarial example in our study. (1) HD: measures the distance between two points clouds in a metric space by computing the nearest original point for each adversarial point and outputting the maximum square distance among all nearest point pairs, as shown below: 
\begin{equation}
    \mathcal{D}_H(X,\hat{X})=\mathop{\min}_{x\in X}\mathop{\max}_{\hat{x}\in \hat{X}} \Vert {x-\hat{x}}\Vert_2^2, 
\end{equation} 
(2) CD: CD is similar to HD but takes an average rather than a maximum, It defined as:
\begin{equation}
    \mathcal{D}_C(X,\hat{X})=\frac{1}{\Vert \hat{X} \Vert_0}\sum_{\hat{x}\in\hat{X}} \mathop{\min}_{x\in X}\Vert{x-\hat{x}\Vert_2^2}.
\end{equation} 
Moreover, for generating adversarial example methods, we use attack success rate to evaluate their effusiveness.
(4) Attack Success Rate (ASR): it computes the attack success rate against defense strategies. For defense strategies, we use defense accuracy (ACC) to evaluate their adversarial robustness. (5) ACC: it measures the accuracy of defense strategies against attack methods.

\subsection{Experimental Results}
\begin{figure}[t]
    \centering
    \includegraphics[width=0.5\textwidth]{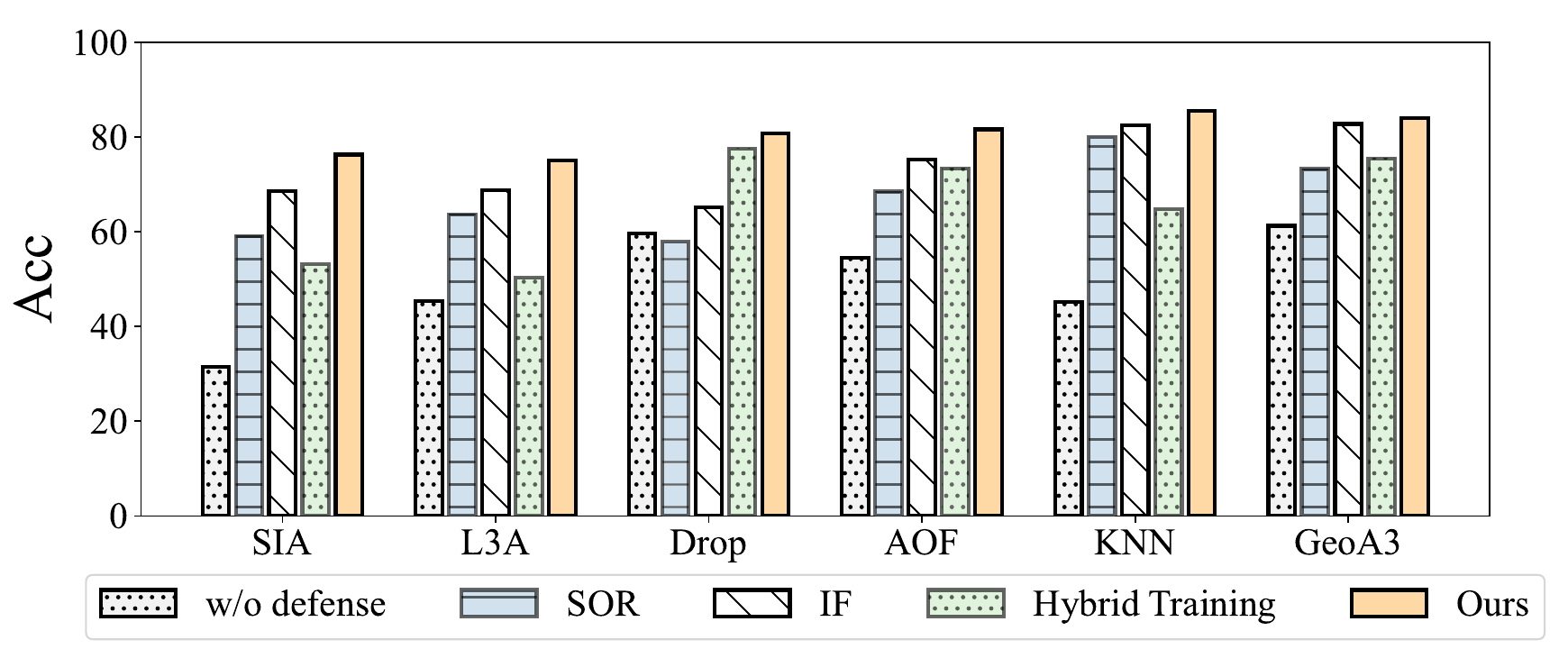}
        \caption{Adversarial robustness of 5 defense strategies under SIA, L3A, Drop, AOF, KNN, and GeoA3 attacks with PointNet.}
    \label{fig:4}
    \vspace{-5pt}
\end{figure}

\begin{figure}[t]
    \centering
    \includegraphics[width=0.5\textwidth]{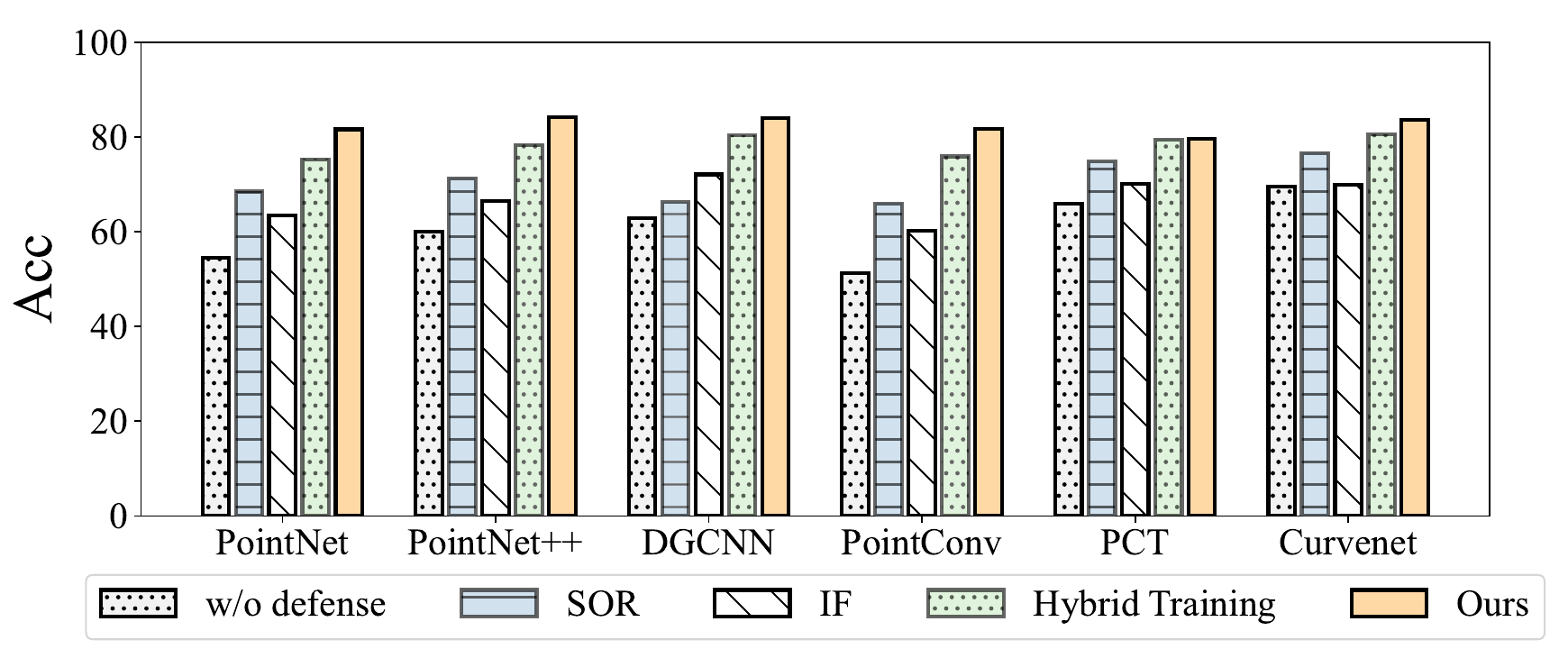}
    \caption{Adversarial robustness of 5 defense strategies under AOF attack with different victim models.}
    \label{fig:5}
    \vspace{-5pt}
\end{figure}

\noindent\textbf{Point Cloud Adversarial Robustness Leaderboad.} Following the process of Figure~\ref{fig:2}, we evaluate the performance of attacks vs. defenses. An illustrated example of leaderboard for point cloud adversarial robustness is presented in Table~\ref{table::summary_of_defense}, where the attacks and defenses are ranked based on their respective average attack success rate and average defense accuracy.

  \noindent\textbf{ 1)} The effectiveness of defense strategies may can vary depending on the models and attacks they are applied to. In Figure~\ref{fig:4} We examine the adversarial robustness of 5 defense strategies across various attacks. Our result reveal that while hybrid training exhibits a high defense accuracy against SIA, Drop, and AOF attacks, it performs poorly against KNN and L3A attacks. In addition, we explore the defense accuracy of defense strategies with different victim models under same attack, as depicted in Figure~\ref{fig:5}. We find that IF-Defense has a large defense performance gap between PointConv and DGCNN. To obtain more convincing results, we recommend that researchers comprehensively evaluate the adversarial robustness of defense strategies by subjecting them to a wide spectrum of attacks and victim models. Such evaluations are essential for accurately evaluating the generalization capabilities of defenses and promoting their practical viability.
    
    \noindent\textbf{2)} Among current point cloud DNNs, it has been observed that models incorporating advanced grouping operations, such as Curve grouping in Curvenet and frequency grouping in GDANet and RPC, exhibit superior performance against various attacks. This performance superiority can potentially be attributed to the high expressiveness of these models' architectures.

    \noindent\textbf{3)} Some defense methods, such as SOR, show worse performance than No defense model. There are two reasons to conduct this phenomenon. For attack, some early attacks (e.g., Pertub and IFGM) exhibit poor transferability. Thus, training settings differences in the target model can degrade the ASR. For defense, some defense methods modifying the shape of point cloud (e.g., SOR and ConvONet) also impacted the classification performance. In some cases, these defensive modifications may degrade the model performances more significantly than the early adversarial attacks, resulting in worse performance than No defense.
    
The complete leaderboard is provided in the appendix. The leaderboard is dynamic and subject to modification with the advent of more potent attacks, defenses, or models. We will analyze the effectiveness, transferability, and imperceptible of adversarial examples in the following.

\noindent\textbf{Attack Effectiveness.}
In Table~\ref{table::summary_of_defense}, we observe the effectiveness of adding points attack is considerably low, indicating that adding point attack poses a significant challenge in affecting the performance of existing models. Furthermore, the average success rate of most shifting points attacks is below 25\%, implying that the majority of existing shifting attacks fail to significantly degrade point cloud DNNs. It indicates most of the previous works may not be applicable in real-world. Therefore, future research should priories designing more practical attack methods that take into account real-world situations.

\noindent\textbf{Attack Transferability and Imperceptibility.} In the benchmark, attackers do not have knowledge about the victim model, which makes the transferability of adversarial examples crucial. To evaluate the transferability of adversarial examples, we selected three wildly-used point cloud DNNs, including PointNet, PointNet++, and DGCNN as the surrogate model. Adversarial examples generated on these surrogate models were tested on all victim models. The transferability results are presented in the appendix. All adversarial examples are tested without any defense strategies, and the transferability was ranked based on the average attack success rate.  Furthermore, we evaluate the imperceptibility of adversarial examples by calculating the Hausdorff distance and Chamfer measurement, respectively. The imperceptibility results are also presented in the appendix. We ranked the adversarial examples based on the average distance of Hausdorff distance and Chamfer measurement. After observing the transferability and imperceptibility results, we identified several good imperceptible adversarial examples, such as GeoA3, IFGM, Perturb, and Add, with poor transferability, indicating a trade-off between imperceptibility and transferability. Therefore, how to balance the transferability and imperceptibility of adversarial examples is a potential research direction.

\subsection{Ablation Study and New Findings}

\begin{figure}[t]
    \centering
    \includegraphics[width=0.5\textwidth]{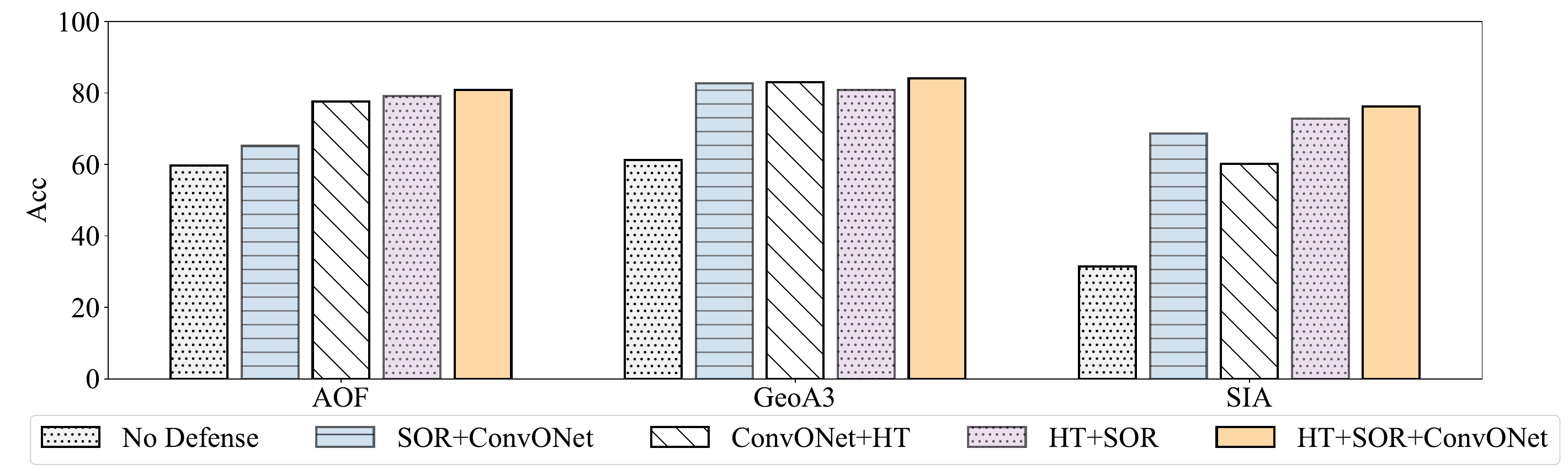}
        \caption{The ablation study of our new defense framework. All attacks are generated on PointNet. HT: hybrid training. }
    \label{fig:6}
\end{figure}

In this section, we present an ablation study of our proposed defense framework, as illustrated in Figure~\ref{fig:6}. Specifically, we conduct experiments by selectively removing individual defense components and evaluating the resulting adversarial robustness against adversarial examples, such as AOF, GeoA3, and SIA. 
From the results, we demonstrate that all modules within our defense framework significantly contribute to the overall robustness of the system. Meanwhile, each module has different effectiveness for robustness. For example, Hybrid training combined with SOR defense can achieve almost the same performance as all modules, but SOR plus ConvOnet gets the lowest defense performance, which reveals the significance of Hybrid training. 

\noindent\textbf{Our New Findings.}
We present new findings on the transferability of adversarial examples in 3D point cloud DNNs. Table~\ref{table::summary_of_defense} and the transferability results in the appendix show that the transferability of point cloud adversarial examples is limited compared with 2D adversarial examples. This limitation can be attributed to the unique characteristics of 3D point cloud DNNs. To enable practical use of adversarial examples in the real-world, it is necessary to design more transferable adversarial examples. Although hybrid training has demonstrated promising accuracy results, it comes with significantly higher training costs. Therefore, investigating novel techniques that can effectively reduce training costs is a potential research direction.

\section{Conclusion and Future Direction}
\label{p_7}
In this paper, we revisit the limitations of previous point cloud adversarial works and establish a comprehensive, rigorous, and unified benchmark for fair comparison of the adversarial robustness of point cloud DNNs. Moreover, we propose a hybrid training method that combines various adversarial examples, including adding, removing, and shifting, to enhance adversarial robustness. Through analysis of the benchmark results, we propose a more robust defense framework by integrating effective defense modules, achieving state-of-the-art adversarial robustness.

The remarkable defense accuracy achieved by ConvOnet demonstrates a direct relationship between the performance of the reconstruction network and the adversarial robustness. Thus, we recommend further investigation and implementation of advanced reconstruction networks to improve adversarial robustness. We highly encourage the community to contribute more advanced point cloud DNNs, attacks, and defenses to enrich future point cloud adversarial robustness benchmarks, benefitting real-world applications.

\textbf{Acknowledgement}
This work was partially supported by the National Science Foundation Grants CRII-2246067, CCF2211163, and the National Natural Science Foundation of China under Grant No. NSFC22FYT45.
\clearpage
{\small
\bibliographystyle{plainnat}
\bibliography{egbib}
}

\end{document}